\pdfoutput=1

\documentclass[11pt]{article}

\usepackage{ACL2023}

\usepackage{times}
\usepackage{latexsym}
\usepackage{graphicx}
\usepackage{amsmath}
\usepackage{amsfonts}
\usepackage{multirow}
\usepackage{booktabs}
\usepackage{stmaryrd}
\usepackage{makecell}
\usepackage{color}
\usepackage{subfigure}
\usepackage{tabularx}

\usepackage[T1]{fontenc}

\usepackage[utf8]{inputenc}

\usepackage{microtype}

\usepackage{inconsolata}

\title{Is ChatGPT Equipped with Emotional Dialogue Capabilities?}

\author{Weixiang Zhao, Yanyan Zhao\thanks{\ \ Corresponding author} \ , Xin Lu, Shilong Wang, Yanpeng Tong, Bing Qin \\
        Research Center for Social Computing and Information Retrieval\\
        Harbin Institute of Technology, China\\
        \texttt{\{wxzhao, yyzhao\}@ir.hit.edu.cn}}

\begin{document}
\maketitle
\begin{abstract} 
This report presents a study on the emotional dialogue capability of ChatGPT, an advanced language model developed by OpenAI. The study evaluates the performance of ChatGPT on emotional dialogue understanding and generation through a series of experiments on several downstream tasks. Our findings indicate that while ChatGPT's performance on emotional dialogue understanding may still lag behind that of supervised models, it exhibits promising results in generating emotional responses. Furthermore, the study suggests potential avenues for future research directions.
\end{abstract}

\section{Introduction}

Emotional dialogue technology is a promising research area that aims to equip chatbots with human-like emotions, enabling them to recognize, understand, and express emotions in their interactions with users, thus generating more engaging and diverse responses. Consequently, emotional dialogue robots have gained significant academic and industrial attention. In recent years, it has emerged as a core technology for enhancing the performance of various application products, including open-domain chatbots \citep{zhou2020design}, intelligent customer service \citep{chen2019antprophet}, and voice assistants \citep{kepuska2018next}. By integrating emotional dialogue technology into these products, chatbots can better understand the needs and emotions of users, providing services that align with user intent. Overall, emotional dialogue technology represents a promising research area with the potential to enhance the performance of various AI applications by enabling robots to interact with users in a more empathetic and human-like manner.

\begin{figure}[htbp]
\centering
\includegraphics[width=0.48\textwidth]{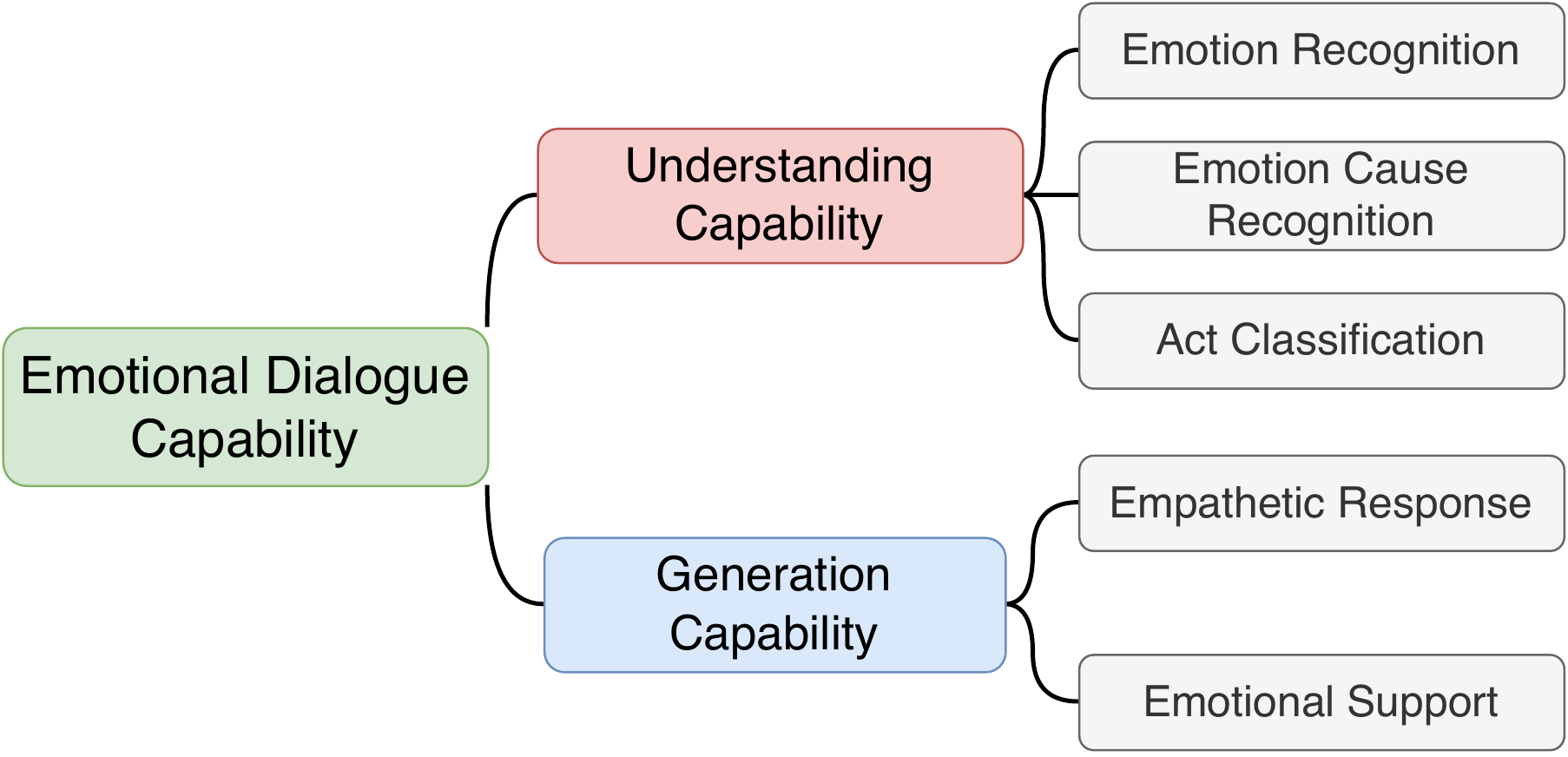}
\caption{The emotional dialogue capability of a chatbot can be divided into two aspects: understanding and generation capability, with several downstream tasks.}
\label{task}
\end{figure}

With the advent of ChatGPT\footnote{https://chat.openai.com}, the field of conversational robots has undergone a revolution. As an advanced large-scale language model, ChatGPT has brought about unprecedented semantic understanding and response generation capabilities for conversational robots, greatly improving the interaction experience with human users. Considering the significant breakthrough of ChatGPT in basic conversational technology, as well as recent research analyzing its performance in various traditional natural language processing tasks \citep{qin2023chatgpt,bang2023multitask,kocon2023chatgpt,wang2023robustness}, whether it exhibits emotional intelligence in dialogue has not yet been explored. Therefore, we are interested in the impact of ChatGPT on the development of emotional dialogue technology. In this report, we will explore the performance of ChatGPT on multiple tasks in the field of emotional dialogue, analyze its strengths and weaknesses, and consider future research directions.

\section{Task Settings}
As shown in Figure \ref{task}, we will compare and analyze the performance of ChatGPT in various downstream tasks based on two dimensions: 
\begin{itemize}
    \item \textbf{Understanding Capability}: Is ChatGPT capable of accurately understanding and interpreting the user's emotions?
    \item \textbf{Generation Capability}: Is ChatGPT capable of eliciting empathy or support towards such emotional states of the user?
\end{itemize}
Detailed definitions and evaluations of each task will be elaborated in the following sections.

\section{Evaluation Methods}
We directly refer to the experimental results of the original papers for state-of-the-art (SOTA) models in each task. For ChatGPT's performance testing, we use the "gpt-3.5-turbo" model of the OpenAI public API (version up to March 8). We evaluated ChatGPT's performance in both zero-shot and few-shot prompting settings for above tasks.

\section{Understanding Capability}
\subsection{Emotion Recognition}
\begin{figure}
	\centering
	\includegraphics[width=0.48\textwidth]{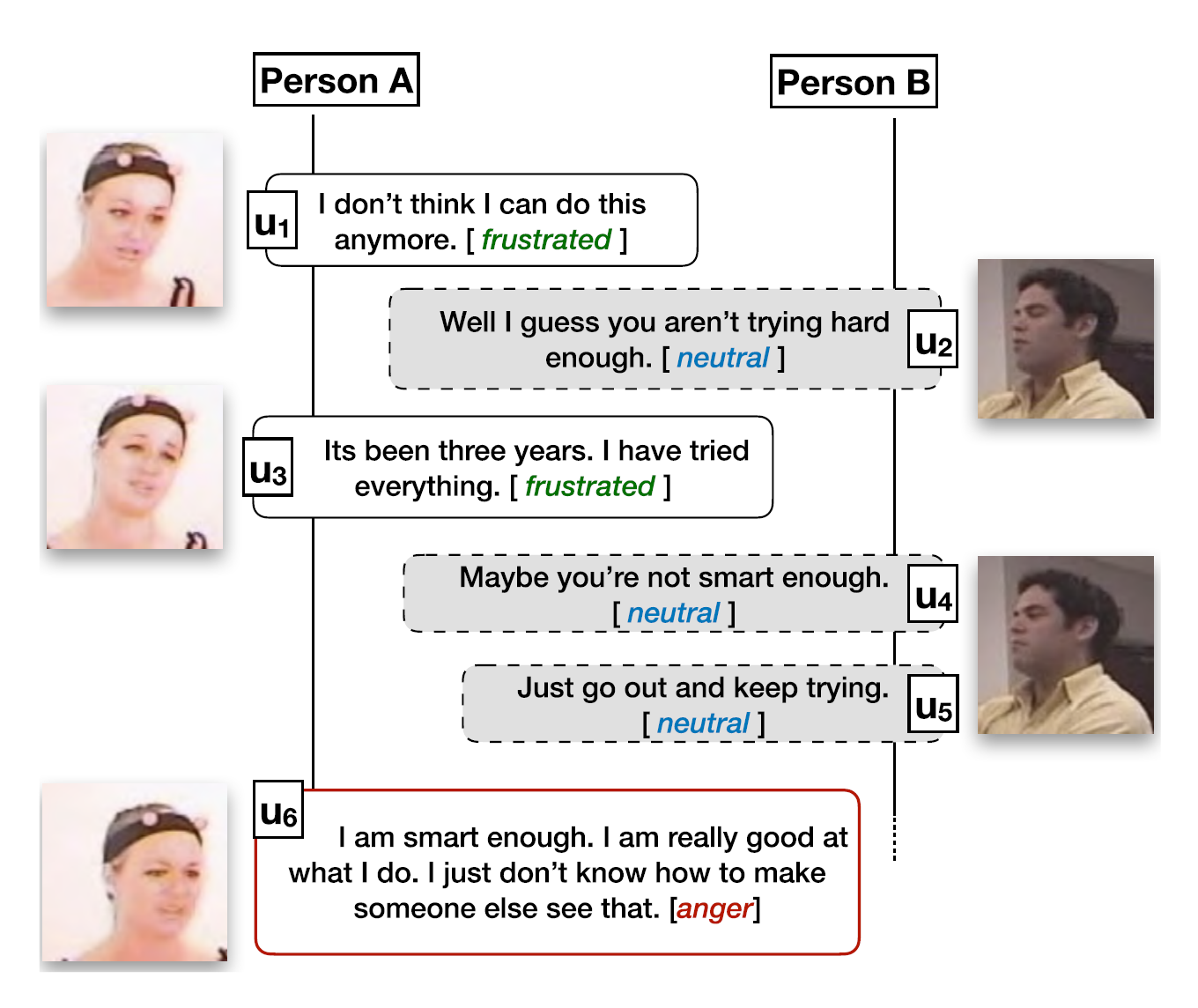}
	\caption{An example \cite{poria2019emotion} of Emotion Recognition in Conversations (ERC).}
	\label{erc_example}
\end{figure}

Emotion Recognition in Conversations (ERC) is a classification task aimed at categorizing emotions within conversational utterances. The input for this task consists of a continuous dialogue, while the output entails the emotional classification of all utterances present. Figure \ref{erc_example} illustrates a straightforward example. Emotion recognition in conversational utterances cannot be reduced to simple single-sentence emotion recognition; it necessitates a holistic examination of the conversation's background, context, and speaker information. 

Emotion Recognition in Conversations (ERC) has extensive applications in a variety of conversational settings, including sentiment analysis of comments on social media platforms and emotional assessment of clients in artificial customer service environments. Furthermore, dialogue emotion recognition can be implemented in chatbots to real-time assess users' emotional states, facilitating the generation of emotionally-driven responses.

\subsubsection{Task Definition}

Given a conversation with several utterances, the goal of Emotion Recognition in Conversations (ERC) is to detect the emotions of all utterances.
Formally, given a conversation $ C = \{ u_1, u_2, ..., u_N \} $ consisting of a sequence of $ N $ utterances, the task is to map the utterance sequence to corresponding emotion label sequence $ Y = \{ y_1, y_2, ..., y_N \} $.

\begin{table*}[htbp]
	\centering
	\normalsize
	\begin{tabular}{ccccc}
		\toprule
		\textbf{Model} & IEMOCAP & MELD & EmoryNLP & DailyDialog\\
		\midrule
            DialogueRNN & 64.76 &63.61 &37.44 &57.32 \\
            IEIN &64.37 &60.72 &- &-\\
            COSMIC &65.28 &65.21 &38.11 &58.48\\
            DialogXL &65.94 &62.41 &34.73 &54.93\\
            DAG-ERC &68.03 &63.56 &39.02 &59.33\\
            DialogueCRN &66.20 &58.39 &- &-\\
            CauAIN &67.61 &65.46 &- &58.21\\
		CoMPM          &69.46 &66.52 &38.93 &\textbf{60.34}\\ 
            MuCDN &- &65.37 &40.09 &-\\
		SPCL           &\textbf{69.74} &\textbf{67.25} &\textbf{40.94} &-\\ 
		\midrule
		ChatGPT 0-shot &44.97 &57.30 &37.47 &40.66 \\
		ChatGPT 1-shot &47.46 &58.63 &35.60 &42.00 \\
		ChatGPT 3-shot &48.58 &58.35 &35.92 &42.39 \\
		\bottomrule
	\end{tabular}
	\caption{Comparison of ChatGPT and other baselines on ERC task.}
	\label{erc_tab}
\end{table*}

\subsubsection{Dataset and Evaluation Metrics}

We evaluate the performance of ChatGPT and all baseline models on four publicly available datasets, IEMOCAP~\cite{DBLP-journals-lre-BussoBLKMKCLN08}, MELD~\cite{poria-etal-2019-meld}, EmoryNLP~\cite{zahiri2017emotion} and DailyDialog~\cite{li2017dailydialog}.

\textbf{IEMOCAP} \ \  The IEMOCAP dataset\footnote{https://sail.usc.edu/iemocap/} was collected by SAIL lab at USC. 
It consists of approximately 12 hours of multimodal conversation data, we only use the text modality in this report. 
The dataset contains 152 dialogues with a total of 7,433 utterances, and it comes with six emotion categories: happy, sad, neutral, anger, excited and frustrated.

\textbf{MELD} \ \ The MELD dataset\footnote{https://affective-meld.github.io/} contains the conversations from Friends TV show transcripts, which is a multimodal extension of the EmotionLines dataset~\cite{DBLP-conf-lrec-HsuCKHK18}. In this report, we only use the text modality. 
The dataset contains 1,433 dialogues with a total of 13,708 utterances, and it comes with seven emotion categories: neutral, surprise, fear, sadness, joy, disgust and anger.

\textbf{EmoryNLP} \ \ The EmoryNLP dataset\footnote{https://github.com/emorynlp/emotion-detection} also contains the conversations from Friends TV show transcripts.
The dataset contains 897 dialogues with a total of 12,606 utterances, and it comes with seven emotion categories: sad, mad, scared, powerful, peaceful, joyful and neutral. 

\textbf{DailyDialog} \ \ The DailyDialog dataset\footnote{http://yanran.li/dailydialog} is a high-quality multi-turn dialogue dataset, with conversations reflecting various topics in daily life. The dataset contains 13,118 dialogues with a total of 102,979 utterances, and it comes with seven emotion categories: neutral, happiness, surprise, sadness, anger, disgust and fear. 

For the IEMOCAP, MELD and EmoryNLP datasets, most papers currently use the Weighted-F1 metric for evaluation; for the DailyDialog dataset, due to the extremely high proportion of neutral utterances, most papers currently use the Micro-F1 metric that excludes the neutral category for evaluation. 

\subsubsection{Main Results}

We selected the current state-of-the-art (SOTA) models: DialogueRNN \citep{majumder2019dialoguernn}, IEIN \citep{lu2020iterative}, COSMIC \citep{ghosal2020cosmic}, DialogXL \citep{shen2021dialogxl}, DAG-ERC \citep{shen2021directed}, DialogueCRN \citep{hu2021dialoguecrn}, CauAIN \citep{zhao2022cauain}, CoMPM \citep{lee-lee-2022-compm}, MuCDN \citep{zhao2022mucdn} and SPCL~\cite{song-etal-2022-supervised} as baseline models and tested the performance of the baseline models and ChatGPT on ERC task.

The experimental results are shown in Table \ref{erc_tab}, from which it can be seen that ChatGPT generally has a performance gap of 3-18 percentage points compared to the most advanced fine-tuned models.

\subsubsection{Case Study}

We show a dialog from the DailyDialogue dataset, which simulates a conversation scenario between a doctor and a patient, as shown in Table \ref{erc_case}.

\begin{table*}
	\centering
	\resizebox{1.0\textwidth}{!}{
	\begin{tabular}{clcc}
		\toprule
		\textbf{Speaker} & \textbf{Dialogue Content} & \textbf{Annotation} & \textbf{Prediction} \\
		\midrule
        A &Good morning. What's the matter with you? &Neutral &Neutral \\
        \midrule
        B &Good morning, doctor. I have a terrible headache. &Neutral &\textcolor{orange}{Sadness} \\
        \midrule
		A &All right, young man. Tell me how it got started. &Neutral &Neutral\\
		\midrule
		B &\makecell[l]{Yesterday I had a runny nose. Now my nose is stuffed up. \\ I have a sore throat. And I’m afraid I've got a temperature. I feel terrible.} &\textcolor{red}{Neutral} &\textcolor{green}{Sadness} \\
		\midrule
		A &\makecell[l]{Don't worry, young man. Lat me give you an examination. \\ First let me take a look at your throat. Open your mouth and say 'ah'.} &Neutral &Neutral \\
		\midrule
		B &Ah. &Neutral &Neutral \\ 
		\midrule
		A &\makecell[l]{Your throat is inflamed. And your tongue is heavily coated. \\ You have all the symptoms of influenza.} &Neutral &\textcolor{orange}{Fear} \\ 
		\midrule
		B &What am I supposed to do then? &Neutral &\textcolor{orange}{Fear} \\
		\midrule
		A &\makecell[l]{A good rest is all you need, and drink more water. \\ I'll write you a prescription.} &Neutral &\textcolor{orange}{Happiness} \\
		\midrule
		B &Thank you very much. &\textcolor{red}{Neutral} &\textcolor{green}{Happiness} \\
		\bottomrule
	\end{tabular}
	}
	\caption{An example of ChatGPT's prediction on ERC task.}
	\label{erc_case}
\end{table*}

\subsubsection{Analysis and Discussion}

In the case study section, we display potential annotation errors in the dataset using red font and instances where ChatGPT rectifies annotation errors in green font. Furthermore, ChatGPT's prediction results include labels in yellow font, signifying an additional issue we identified: an inconsistency between ChatGPT and the dataset's guidelines. Examining these actual prediction samples reveals that the primary challenge for ChatGPT is the deviation between its criteria and those of the dataset. While the dataset annotation likely adhered to specific guidelines to determine the corresponding emotion for a given situation, ChatGPT operates under its own interpretation and standards. For instance, in the dialogue between the physician and patient, when the patient describes their headache symptoms, the dataset annotation classifies the emotion as neutral, while ChatGPT deems it sadness. This discrepancy cannot be attributed to one being correct and the other incorrect but rather highlights the differing standards employed.

Upon further discussion, this misalignment of standards may not stem from ChatGPT's capabilities but could be attributed to the few-shot prompting setting. As annotation guidelines become increasingly intricate and involved, it becomes implausible to encompass them with merely a handful of examples, which is an inherent constraint of few-shot prompting.

This insight allows for conjecture on possible future directions: if the objective is not to strictly conform to specific guidelines, then enhancements based on few-shot prompting settings, such as ChatGPT, are viable. However, utilizing dataset labels for evaluation may be unsuitable, potentially necessitating extensive human evaluations. Conversely, if the goal is to strictly adhere to specific guidelines, few-shot prompting settings may not be the optimal choice, with supervised fine-tuned models remaining the superior alternative.

\begin{figure}
\centering
\includegraphics[width=0.48\textwidth]{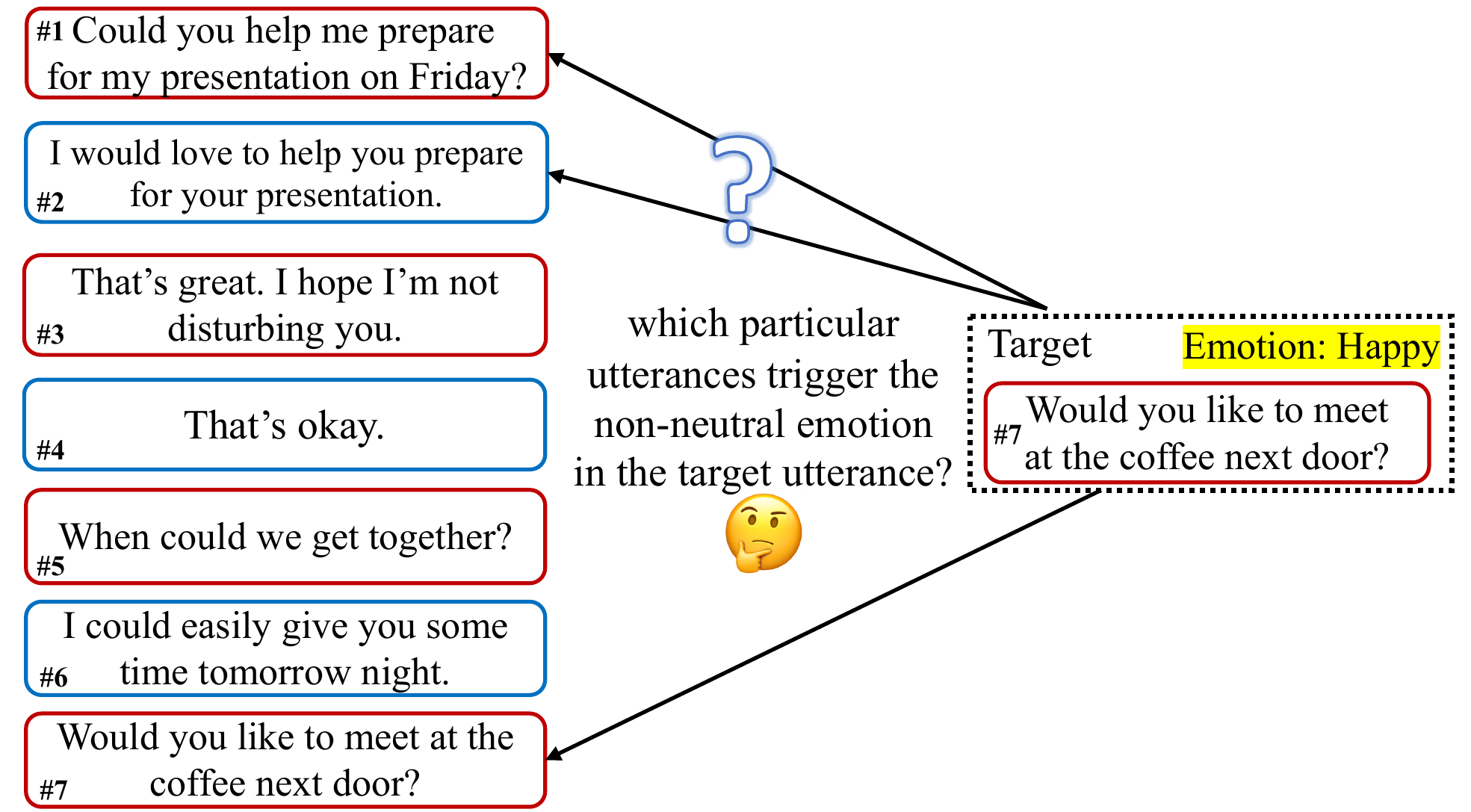}
\caption{An example from RECCON-DD dataset \cite{poria2021recognizing} for identifying the causal utterances.}
\label{cee_example}
\end{figure}

\subsection{Emotion Cause Recognition}
Step further on ERC, recognizing emotion cause in conversations serves to fully understand the emotional situations of the user and is beneficial to improve interpretability and performance in affect-based models. And \citet{poria2021recognizing} introduce a new task named RECCON with annotated emotion causes in conversations. It includes two different sub-tasks: Causal Span Extraction (CSE) and Causal Emotion Entailment (CEE). We focus on the CEE sub-task in this report and its goal is to predict which particular utterances in the conversation history trigger the non-neutral emotion in the target utterance, as illustrated in Figure \ref{cee_example}.

\subsubsection{Task Definition}
First, we define the problem of the CEE task. Given a conversation that consists of $t$ consecutive utterances $\left\{u_1, u_2, \cdots, u_t \right\}$ with the corresponding emotion label $\left\{e_1, e_2, \cdots, e_t \right\}$ between two speakers, the goal of this task is to predict which particular utterances $u_i$ ($i \le t$) in the conversational history are responsible for the non-neutral emotion $e_t$ in the target utterance $u_t$. And $u_i$ is a positive example if it contains the cause of non-neutral emotion in the target utterance and a negative example otherwise.

\subsubsection{Dataset and Evaluation Metrics}
In this study, we evaluate the performance of ChatGPT and all baseline models on the widely-used benchmark dataset RECCON-DD\footnote{https://github.com/declare-lab/RECCON.}, which features utterance-level emotion labels and emotion cause labels annotated by \citet{poria2021recognizing}, based on the popular dialogue dataset DailyDialog \citep{li2017dailydialog}. Specifically, we focus on the causal pairs extracted from the dialogue history, filtering out any repetitive instances. 

To be consistent with all baseline methods, we report the F1 scores of both negative and positive causal pairs and the macro F1 scores of them.

\begin{table}
\normalsize
\centering
\resizebox{\linewidth}{!}{
\begin{tabular}{cccc}
\toprule
\textbf{Model} & Neg. F1 & Pos. F1 & macro F1\\
\midrule
RoBERTa-Base &88.74 &64.28 &76.51\\ 
RoBERTa-Large &87.89 &66.23 &77.06\\
KEC &88.85 &66.55 &77.70\\ 
KBCIN &89.65 &68.59 &79.12\\
TSAM &\textbf{90.48} &\textbf{70.00} &\textbf{80.24}\\
\midrule
ChatGPT 0-shot &85.25 &51.33 &68.29 \\
ChatGPT 1-shot &82.10 &52.84 &67.47 \\
\bottomrule
\end{tabular}
}
\caption{Comparison of ChatGPT and other baselines on CEE task.}
\label{cee_tab}
\end{table}

\subsubsection{Main Results}
We compare the performance of ChatGPT 0-shot and 1-shot ability with that of supervised baseline models. To be more specific, we select the pretrained-model based methods \citep{poria2021recognizing, zhang2022tsam} and commonsense-based methods \citep{li2022neutral,zhao2022knowledge}.

Results are presented in Table \ref{cee_tab}. On one hand, ChatGPT achieve comparable performance on the recognition of negative causal pairs. On the other hand, it still holds a 11.95\% gap in terms of macro F1 score compared to the SOTA supervised method.

\subsubsection{Analysis and Discussion}
Through an analysis of error cases in ChatGPT, it has been identified that the primary cause of the performance gap between ChatGPT's Pos. F1 score and the state-of-the-art (SOTA) lies in the presence of numerous emotionally charged samples within the target sentence itself. ChatGPT has shown a tendency to overlook these types of examples and instead focuses on identifying causal statements from the conversation context. This observation aligns with the previously discussed analysis of emotion recognition in dialogues. One of the primary reasons for ChatGPT's poor performance is the considerable discrepancy between its prediction standard and the dataset's annotation standard. Additionally, ChatGPT's performance decreases when given a specific example, highlighting the importance of a thorough understanding of the complex annotation standards for tasks such as identifying the causes of emotions. Unlocking the full potential of ChatGPT's performance requires a deep understanding of the dataset's specifications, allowing it to overcome its own language model prior and achieve results that better align with downstream testing data.

\subsection{Dialog Act Classification}

Dialogue Act Classification (DAC) is a classification task that involves labeling the dialogue acts for each utterance in a dialogue. The input for this task is a continuous segment of dialogue consisting of several utterances, and the output is the labeling of dialogue acts for all utterances. It is assumed that each round in a dialogue involves a particular dialogue act, and therefore, the label set does not include a neutral label. An example of DAC is shown in Table \ref{dac_case}. Similar to ERC, DAC is not a simple sentence classification problem; instead, it requires considering the context of the conversation.

Categorizing dialogue acts enables a better understanding of a conversation's intentions, emotions, and context. This task has extensive applications, including dialogue systems, customer service chatbots, smart speakers, and sentiment analysis. Dialogue act classification in dialogue systems and customer service chatbots can improve robots' comprehension of users' needs and intentions, resulting in better service and responses. In sentiment analysis, dialogue action classification can facilitate an improved understanding of dialogue behaviors under various emotional states, allowing for a more thorough analysis and comprehension of users' emotional states.

\begin{table*}
	\centering
	\resizebox{1.0\textwidth}{!}{
	\begin{tabular}{clcc}
		\toprule
		\textbf{Speaker} & \textbf{Dialogue Content} & \textbf{Annotation} & \textbf{Prediction} \\
		\midrule
        A &When can we expect you for dinner? Can you come tonight? &\textcolor{red}{Directive} &\textcolor{orange}{Question} \\
        \midrule
        B &Not tonight. I promised to go to a concert with my sister. &\textcolor{red}{Commissive} &\textcolor{orange}{Inform} \\
        \midrule
	A &Well ... How about Friday then? &\textcolor{red}{Directive} &\textcolor{orange}{Question}\\
	\midrule
	B &That sounds fine. &Commissive &Commissive \\ 
		\bottomrule
	\end{tabular}
	}
	\caption{An example of ChatGPT prediction on DAC task.}
	\label{dac_case}
\end{table*}

\begin{table}
	\centering
	\begin{tabular}{lcc}
		\toprule
		\textbf{Model} & \textbf{Acc} & \textbf{Weighted-F1} \\
		\midrule
  	    CASA          &- &0.78 \\
            DCR-Net + Co-Attention &- &0.79 \\  
            Co-GAT          &- &0.79 \\
            WEAKDAP          &0.84 &- \\  
		\midrule
		ChatGPT 1-shot &0.67 &0.65 \\
		ChatGPT 1-shot + P.E &0.71 &0.70\\
		ChatGPT 3-shot &0.73 &0.71\\
  		ChatGPT 3-shot + P.E &0.73 &0.72\\
		\bottomrule
	\end{tabular}
	\caption{Comparison of ChatGPT and other baselines on DAC task. Experimental results are partly cited from Co-GAT\citep{qin2021co}. P.E stands for prompt engineering.}
	\label{dac_tab}
\end{table}

\subsubsection{Task Definition}

Given a conversation with several utterances, the goal of DAC is to detect the dialog acts of all utterances.
Formally, given a conversation $ C = \{ u_1, u_2, ..., u_N \} $ consisting of a sequence of $ N $ utterances, the task is to map the utterance sequence to corresponding dialog act label sequence $ Z = \{ z_1, z_2, ..., z_N \} $.

\subsubsection{Dataset and Evaluation Metrics}

The experiments were carried out on the DailyDialog~\cite{li2017dailydialog} dataset, as previously introduced in the ERC section. This study solely employed the act tags available in the dataset. To evaluate the classification task, both weighted-F1 and macro-F1 were used as performance metrics. However, ChatGPT assigned meaningless labels beyond the four categories, which greatly affected the macro-average value. To address this issue, the evaluation metric used in this task was the weighted average F1 score.

\subsubsection{Main Results}

We select the current state-of-the-art (SOTA) model \textbf{WEAKDAP} \citep{chen2022weakly} and some other strong baselines \textbf{CASA} \citep{raheja2019dialogue}, \textbf{DCR-Net} \citep{qin2020dcr} and \textbf{Co-GAT} \citep{qin2021co}. Then we teste the performance of the baseline models and ChatGPT on Dialog Act Classification(DAC) task. 

The experimental results are shown in Table \ref{dac_tab}, from which it can be seen that ChatGPT generally has a performance gap of 11-17 percentage points compared to the baseline fine-tuned models.

\subsubsection{Case Study}

We show a dialog from the DailyDialogue dataset, which simulates a conversation scenario between a doctor and a patient, as shown in Table \ref{dac_case}.

\subsubsection{Analysis and Discussion}

ChatGPT has demonstrated limited understanding of labels such as directives and promises. Specifically, it tends to conflate question with directive and inform with commissive, as illustrated in the case study. Due to the lack of clear definitions, these two labels have a semantic overlap, making it difficult to distinguish between them. For example, "Can you come today?" is a directive question, while "I promised to go to the concert with my sister" is a commissive inform. It should be noted that this issue does not necessarily indicate ChatGPT's poor understanding of conversational actions, but rather highlights a disparity between the model's labeling system and that of the dataset. To address this challenge, incorporating comprehensive label explanations into prompts, referred to as prompt engineering, can significantly enhance evaluation metrics.

The experimental results demonstrate that the few-shot setting is the most effective prompt enhancement method for ChatGPT in this task. The approach does not require complex prompt engineering for the commissive and directive labels, yet it can significantly improve the evaluation metrics. The experiment employed three samples for few-shot and a simple prompt engineering design. We have reasons to believe that carefully selecting more examples and refining the prompt engineering can further decrease the difference between ChatGPT's label system and the original label system of the dataset, leading to an improvement in ChatGPT's performance in this task. Nevertheless, as mentioned in the previous two tasks, the evaluation system's alignment with the dataset label system warrants further consideration.

The results of the experiment indicate that prompt engineering can enhance ChatGPT's performance in a specific task. However, since the prompts used in this study were not intricate, the upper limit of ChatGPT's performance was not tested. We believe that a more sophisticated prompt engineering approach could potentially improve ChatGPT's performance on this task. Developing more customized and high-quality prompts that are well-suited to the task and maximizing the potential of large models may represent a new paradigm for task-solving.

\begin{table*}
\normalsize
\centering
\begin{tabular}{cccccccc}
\toprule
\textbf{Model} & \textbf{D-1} & \textbf{D-2} & \textbf{B-1} & \textbf{B-2} & \textbf{B-3} & \textbf{B-4} & \textbf{R-L} \\
\midrule
Multi-TRS &0.44 &1.98 &21.42 &7.60 &3.67 &2.13 &21.66\\
MoEL  &0.58 &2.91 &21.70 &7.75 &3.58 &1.96 &\textbf{22.08}\\
MIME &0.42 &1.71 &\textbf{22.20} &\textbf{8.07} &3.89 &2.22 &21.54\\
EmpDG &0.44 &1.91 &21.99 &8.02 &3.86 &2.19 &22.03\\
CEM &0.65 &3.03 &18.69 &6.84 &3.37 &1.92 &21.65\\
EmpSOA &0.69 &3.87 &21.41 &8.06 &\textbf{4.14} &\textbf{2.41} &21.64\\
SEEK &0.66 &2.74 &15.21 &4.40 &1.81 &0.85 &19.44\\
\midrule
ChatGPT 2-shot &\textbf{4.17} &\textbf{23.23} &12.99 &4.07 &1.90 &1.05 &12.70 \\
\bottomrule& 
\end{tabular}
\caption{Comparison of ChatGPT against state-of-the-art baselines in terms of the automatic evaluation. The best results among all models are highlighted in bold.}
\label{emp_auto_tab}
\end{table*}

\begin{figure}[htbp]
\centering
\includegraphics[width=0.48\textwidth]{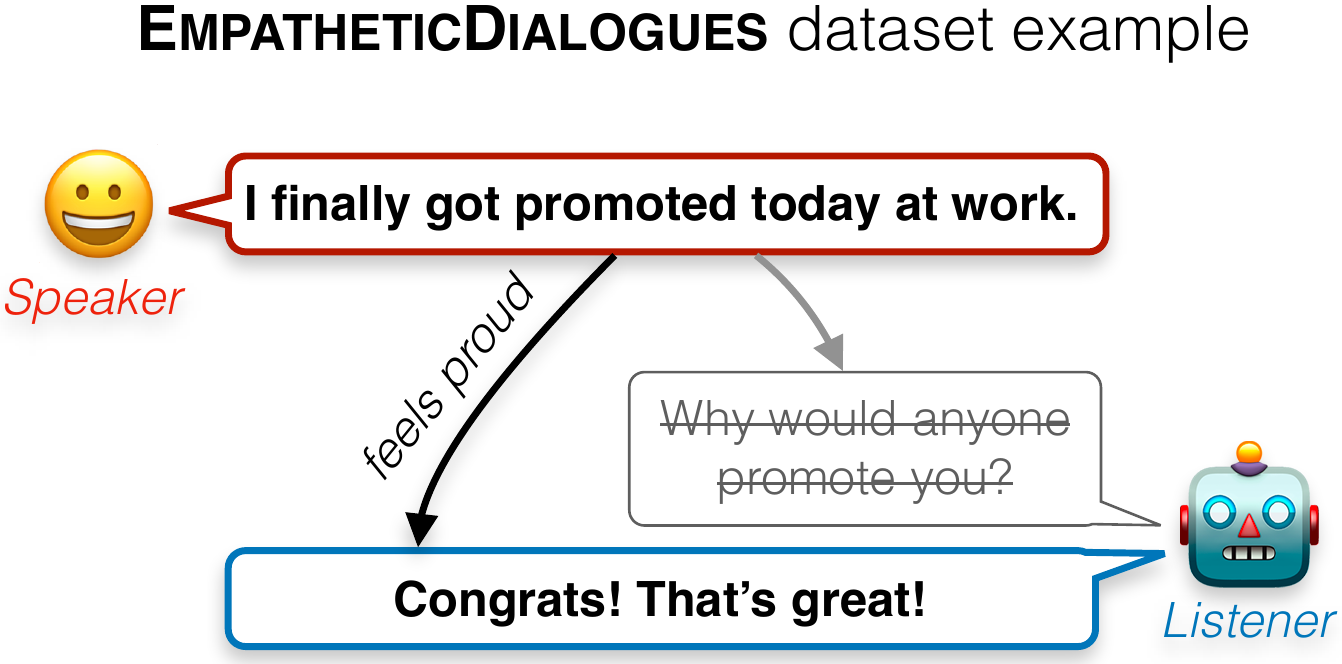}
\caption{An example of empathetic conversation from \textsc{EmpatheticDialogue} dataset \cite{rashkin2018towards}.}
\label{emp_example}
\end{figure}

\section{Generation Capability}

\subsection{Empathetic Response Generation}
\label{empathy}
Empathy is a highly valued characteristic in engaging human conversation, and it is also a crucial element in the development of human-like chatbots. We concentrate on the task of generating empathetic responses \citep{rashkin2018towards} that accurately comprehend the user's emotions and circumstances and provide appropriate feedback. An example is shown in Figure \ref{emp_example}.

\subsubsection{Task Definition}
We formally define the task of empathetic response generation. Specifically, we consider a dialogue history denoted by $D=[X_1, X_2, \cdots, X_N]$, where $N$ represents the number of utterances exchanged between the user and the system. Each utterance $X_i = [w^i_1, w^i_2, \cdots, w^i_m]$ consists of a sequence of $m$ words. Additionally, the conversation is annotated with an emotion label $e$ from a set of 32 available emotions, indicating the emotional tone of the user. The objective of this task is to generate a coherent response $Y$ from the perspective of the system that not only takes into account the dialogue history $D$ but also demonstrates empathy towards the user's situation and emotions.

\subsubsection{Dataset and Evaluation Metrics}
We conduct our experiments on \textsc{EmpatheticDialogues}\footnote{https://github.com/facebookresearch/EmpatheticDialogues} dataset \citep{rashkin2018towards}, which is a large-scale, multi-turn corpus consisting of 25k empathetic conversations, collected via Amazon Mechanical Turk. It features conversations between a speaker and a listener. Notably, the dataset includes 32 emotion labels, evenly distributed to indicate the personal emotional states of the user.

\begin{table}
\normalsize
\centering
\begin{tabular}{llcc}
\toprule
\multirow{2}{*}{\textbf{ChatGPT vs.}} & \multicolumn{3}{c}{\textbf{EmpSOA}}  \\
\cmidrule(r){2-4}
    & Win  & Lose  & Tie \\
    
\midrule
Coherence                & \textbf{49.00}   & 6.33            & 44.67   \\
Empathy               & \textbf{54.33}   & 8.00            & 37.67     \\
Informativeness               & \textbf{78.67}   & 6.00             & 15.33       \\
\bottomrule
\end{tabular}
\caption{The results of the human evaluation.}
\label{emp_human_tab}
\end{table}

\begin{table*}
\small
\centering
\begin{tabularx}{\textwidth}{c|l}
\toprule
\textbf{Emotion} & Devastated \\
\textbf{Context} & My dog I had for five years just passed away. He was hit by a car last night.\\
\midrule
\textbf{MIME} & Oh no! I am so sorry to hear that. \\
\textbf{EmpDG} & Oh no! I am so sorry to hear that.\\
\textbf{CEM} & Oh no! I am so sorry to hear that. \\
\textbf{EmpSOA} & Oh no! I am so sorry to hear that. \\
\midrule
\textbf{ChatGPT} & I'm so sorry for your loss. Losing a beloved pet can be incredibly hard. Would you like to talk about it more?\\
\midrule
\textbf{Ground-Truth} & Oh I am so sorry. That must be fresh on your heart, may he rest in peace.\\
\bottomrule
\end{tabularx}
\vspace{0.2cm}
\begin{tabularx}{\textwidth}{c|l}
\toprule
\textbf{Emotion} & Content \\
\multirow{3}{*}{\textbf{Context}} & My wife made me pancakes for breakfast. I have a full belly and feel rather happy now.\\
& Yum! I love pancakes. Is it your favorite food too? \\
& One of my favorite breakfast foods. She also made some thick sliced bacon. She treats me so well. \\
\midrule
\textbf{MIME} & That is a great idea. I love my girlfriend too. \\
\textbf{EmpDG} & That is a great attitude to have!\\
\textbf{CEM} & That is good, I love it! \\
\textbf{EmpSOA} & That sounds so sweet. I am glad you have a great time! \\
\midrule
\textbf{ChatGPT} & It's so wonderful to have someone who cares for and treats us well. Enjoy your delicious breakfast!\\
\midrule
\textbf{Ground-Truth} & That is great you have a sweet wife!\\
\bottomrule
\end{tabularx}
\caption{Case study of the generated empathetic responses by ChatGPT and the baselines.}
\label{emp_case}
\end{table*}

\paragraph{Automatic Evaluation.} We apply two kinds of automatic metrics for evaluation: (1) BLEU-n (\textbf{B-1}, \textbf{B-2}, \textbf{B-3}, \textbf{B-4}) \citep{papineni2002bleu} and ROUGE-L (\textbf{R-L}) \citep{lin2004rouge} evaluate the lexical and semantic aspects of the generated responses; (2) Distinct-$n$ (\textbf{Dist}-$n$) \citep{li2015diversity} evaluates the diversity of the generated responses by measuring the ratio of unique $n$-grams.

\paragraph{Human Evaluation.} We conduct an aspect-based pairwise preference test to evaluate the quality of responses generated by ChatGPT and EmpSOA. Specifically, we randomly sample 100 pairs of responses and ask three professional annotators to assess which response is better based on three criteria: \textbf{Coherence}, \textbf{Empathy}, and \textbf{Informativeness}. The Coherence criterion assesses the coherence and relevance of the response to the dialogue history, while the Empathy criterion evaluates the degree of empathy displayed towards the user's feelings and situations. Finally, the Informativeness criterion gauges the amount of information related to the dialogue history contained in the response.

\subsubsection{Main Results}
We compare ChatGPT with the following competitive baselines: \textbf{Multi-TRS} \citep{rashkin2018towards}, \textbf{MoEL} \citep{lin2019moel}, \textbf{MIME} \citep{majumder2020mime}, \textbf{EmpDG} \citep{li2019empdg}, \textbf{CEM} \citep{sabour2022cem}, \textbf{EmpSOA} \citep{zhao2022don} and \textbf{SEEK} \citep{wang2022empathetic}.

The experimental results presented in Table \ref{emp_auto_tab} demonstrate that the responses generated by ChatGPT exhibit greater diversity, with an average length of 21.77 tokens, compared to the baseline methods, which have an average length of 14.89 tokens. It is important to note, however, that this increased diversity may lead to a higher degree of mismatch with the golden, indicating a potential trade-off between response diversity and accuracy.

\subsubsection{Analysis and Discussion}
During empathetic response generation, ChatGPT tends to produce longer and more diverse responses, particularly when the user is in a negative emotional state. ChatGPT tends to suggest solutions to address the problems users are facing, leading to a deviation from real responses. This is also the reason why ChatGPT performs significantly worse than the baseline method on word overlap-based automatic evaluation metrics. Furthermore, based on human evaluation, the baseline method's coherence and empathy ability are comparable to ChatGPT, but the amount of information provided in the responses differs significantly. ChatGPT's generated responses are able to fully understand the user's situation, expand on the user's topic, and provide more effective information to the user. As shown in the cases in Table \ref{emp_case}, regardless of whether the user's emotion is positive or negative, ChatGPT's responses are more specific to the context of the conversation, rather than generating generic responses. However, in terms of empathy expression, ChatGPT often repeats the pattern of restating the user's emotion before expanding on the information, which can make the user feel bored.

In light of future directions for this task, several considerations emerge. First and foremost, investing in expanding the parameter and data volume of dialogue models has demonstrated its effectiveness in improving performance, as exemplified by ChatGPT. Second, enhancing the model's ability to empathize on a personalized level remains crucial, as it is apparent that relying on template-like expressions for empathy does not accurately align with authentic human empathetic conversation. Lastly, differences in model performance revealed through automatic and human evaluation underscore the current lack of a suitable evaluation metric for assessing the quality of empathetic dialogue systems.

\begin{figure}
\centering
\includegraphics[width=0.48\textwidth]{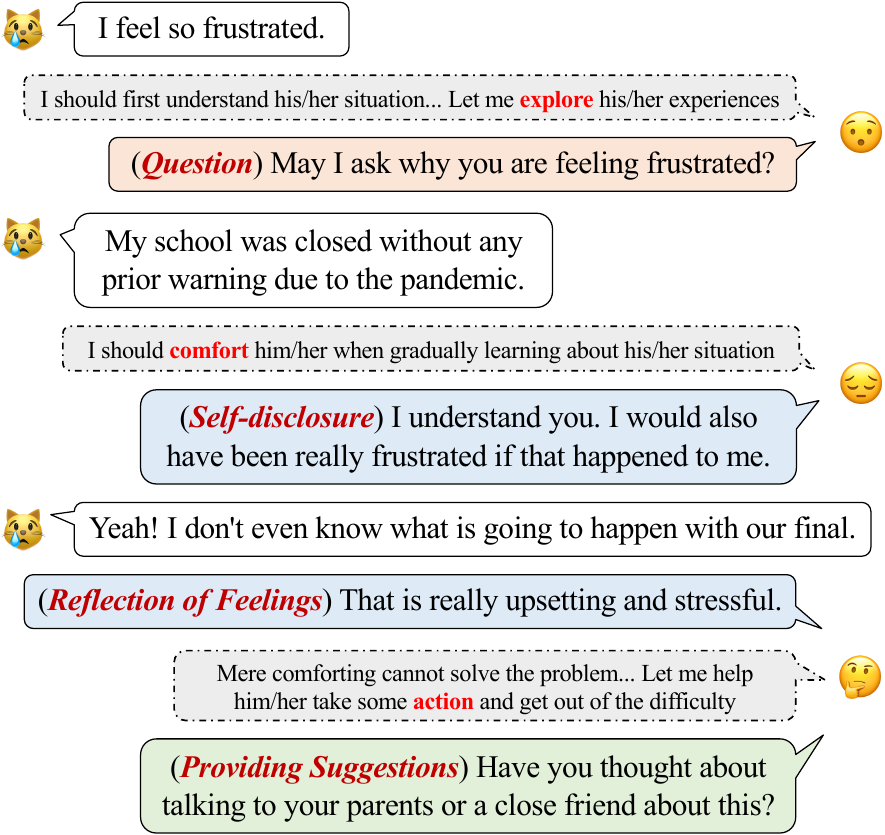}
\caption{An example of emotional support conversation from \textsc{ESConv} dataset \cite{liu2021towards}.}
\label{esc_example}
\end{figure}

\begin{table*}
\normalsize
\centering
\begin{tabular}{ccccccccc}
\toprule
\textbf{Model} & \textbf{Acc} & \textbf{D-1} & \textbf{D-2} & \textbf{B-1} & \textbf{B-2} & \textbf{B-3} & \textbf{B-4} & \textbf{R-L} \\
\midrule
BlenderBot-Joint &17.69 &2.96 &17.87 &18.78 &7.02 &3.20 &1.63 &14.92\\
MISC &31.67 &4.62 &20.17 &16.31 &6.57 &3.26 &1.83 &17.24 \\
GLHG &- &3.50 &21.61 &19.66 &7.57 &3.74 &2.13 &16.37\\
MultiESC &\textbf{42.01} &- &- &\textbf{21.65} &\textbf{9.18} &\textbf{4.99} & \textbf{3.09} &\textbf{20.41}\\
\midrule
ChatGPT 1-shot &17.0 &\textbf{5.92} &\textbf{31.38} &13.91 &4.53 &1.96 &1.02 &13.19 \\
\bottomrule& 
\end{tabular}
\caption{Comparison of ChatGPT against state-of-the-art baselines in terms of the automatic evaluation on ESC task. The best results among all models are highlighted in bold.}
\label{esc_auto_tab}
\end{table*}

\begin{table}
\normalsize
\centering
\begin{tabular}{llcc}
\toprule
\multirow{2}{*}{\textbf{ChatGPT vs.}} & \multicolumn{3}{c}{\textbf{MISC}}  \\
\cmidrule(r){2-4}
    & Win  & Lose  & Tie \\
    
\midrule
Fluency                & \textbf{61}   & 6            & 22   \\
Identification           & \textbf{68}        & 6    & 15     \\
Empathy               & 16   & \textbf{40}             & 33     \\
Suggestion               & \textbf{73}   & 3             & 13       \\
\midrule
Overall                  &\textbf{65}   & 12             & 12     \\
\bottomrule
\end{tabular}
\caption{The results of the human interaction evaluation between ChatGPT and MISC.}
\label{esc_human_tab}
\end{table}

\subsection{Emotional Support Conversation}

Emotional support conversation (ESC) is a dialogue response generation task that aims to provide assistance to seekers who approach for help while experiencing negative emotions. The task requires input of the dialogue history between the seeker and supporter, and outputs an emotional support response from the supporter. An example is shown in Figure \ref{esc_example}. It consists of three stages: first, the supporter needs to \textcolor[RGB]{239, 134, 51}{explore} the situations and identify the issue the seeker is facing; second, the supporter should offer \textcolor[RGB]{111, 164, 211}{comfort} to the seeker, and finally, take \textcolor[RGB]{92, 142, 79}{action} to provide advice or information to help the seekers address their problem. The supporter may utilize eight distinct strategies, including questioning, restatement or paraphrasing, reflection of feelings, self-disclosure, affirmation and reassurance, providing suggestions, information, and others. Please refer to Appendix \ref{app:stra} for the detailed definitions of these strategies.

\subsubsection{Task Definition}
Formally, let $D=[X_1, X_2, \cdots, X_N]$ denote a dialogue history consisting of $N$ utterances between a seeker and a supporter. Each utterance $X_i = [w^i_1, w^i_2 \cdots, w^i_m]$ is a sequence of $m$ words, and utterances from the supporter's turns are associated with the support strategy $S_i$. Our objective is to generate the next coherent and supportive utterance $Y$ from the supporter's perspective, with the aim of alleviating the seeker's distress.

\subsubsection{Dataset and Evaluation Metrics}
Our experiments are conducted on the \textsc{ESConv}\footnote{https://github.com/thu-coai/Emotional-Support-Conversation.} dataset \citep{liu2021towards}. It is an English dataset. To construct the dataset, they recruited crowd-workers, who had learned the common procedures and strategies for providing emotional support, to converse with volunteers that needed emotion support through an online platform. It contains
1,300 long dialogues with 38,350 utterances. There is an average of 29.5 utterances per dialogue and an average 16.7 tokens per utterance.

\paragraph{Automatic Evaluation.} We employ the same automatic metrics as those used in the empathetic response generation task. In addition, Accuracy (\textbf{Acc}) of the strategy prediction is utilised to evaluate the capability to choose the supportive strategy.

\paragraph{Human Evaluation.} Following \citet{liu2021towards}, we recruit one professional annotators to interact with the models for human evaluation. Specifically, we recruit a professional annotator to evaluate 89 dialogues randomly sampled from the test set of \textsc{ESConv}. The annotator assumes the role of a seeker and engages in conversations with both models, and evaluates them based on five criteria: (1) \textbf{Fluency}: the coherence and smoothness of the generated responses, (2) \textbf{Identification}: how effectively the models identify and address the seeker's problems, (3) \textbf{Empathy}: the level of empathetic understanding displayed by the models towards the seeker's feelings and situation, (4) \textbf{Suggestion}: the quality of the suggestions offered by the models, and (5) \textbf{Overall}: the overall effectiveness of the models in providing emotional support.

\subsubsection{Main Results}
We compare ChatGPT with the following competitive baselines on ESC task: \textbf{BlenderBot-Joint} \citep{liu2021towards}, \textbf{MISC} \citep{tu2022misc}, \textbf{GLHG} \citep{peng2022control} and \textbf{MultiESC} \citep{cheng2022improving}. They are based on generative pretrained model BlenderBot \citep{roller2020recipes} and BART \citep{lewis2019bart}.

\subsubsection{Analysis and Discussion}
As shown in Table \ref{esc_auto_tab}, the responses generated by ChatGPT exhibits both long and diverse characteristics, leading to superior performance over SOTA methods in terms of the automatic evaluation metric Distinct-n. However, such diversity can also introduce deviations from golden responses.

And for human evaluation displayed in Table \ref{esc_human_tab}, since one of the characteristics of ESC is to provide users with suggestions and effective information to help them get out of their dilemma, this happens to be consistent with the generation preference of ChatGPT of generating informative responses. Therefore it demonstrates excellent performance in this task. However, in terms of empathy, the reason why SOTA methods outperform ChatGPT is that ChatGPT is too eager to give corresponding advice and coping strategies once it confirms the dilemma the user faces, ignoring the comfort and care of the user's emotions. But this does not imply that ChatGPT lacks the ability of eliciting empathy. Its excellent performance in empathetic response generation tasks can prove that it can comfort users empathetically (Section \ref{empathy}). Through appropriate prompt engineering, we believe that ChatGPT can "slow down" and carry out sufficient emotional guidance before giving users advice. Compared with MISC, ChatGPT can display more diverse and effective supportive responses. However, MISC cannot learn this point from existing datasets because real advice in the corpus itself is limited.

Future research on ESC should focus on how to make the model adaptively control the rhythm of emotion support to avoid giving advice too hastily or repeating ineffective comfort. Furthermore, exploring more reasonable automatic evaluation metrics that align with human evaluation is a research direction that deserves further exploration.

\section{Conclusion}

In this report, we conduct a preliminary exploration of the emotional conversation capability of ChatGPT. It should be noted that our experimental results may not fully reflect the optimal performance of ChatGPT in the corresponding task. With more refined prompt engineering and context example selection, we believe that the performance of ChatGPT can be further improved. One of the future directions for emotional dialogue understanding is to explore the alignment of ChatGPT with labeling standards. As for emotional dialogue generation, it is important to investigate reasonable automatic evaluation metrics to measure model performance, as the results obtained from widely used automatic and human evaluations may differ.

\section*{Limitations}

\paragraph{Limitations of Model Selection.} Due to resource constraints, we have limited our evaluation to a single representative language model (LLM), namely the gpt-3.5-turbo variant of ChatGPT. However, the field of LLMs is rapidly advancing, with numerous other notable models such as the GPT-3.5 series (including text-davinci-002, code-davinci-002, textdavinci-003), as well as the recently announced GPT-4. It is our belief that a comprehensive analysis of the abilities of various LLMs in the emotion dialogue capability will be necessary in the future.

\paragraph{Limitations of Automic Evaluation.} Owing to limited resources, we only employ simple prompt engineering and conduct few-shot prompting under a low-resource setting (no more than 3-shot). However, this approach may not accurately reflect the optimal performance of ChatGPT on the corresponding downstream tasks.

\paragraph{Limitations of Human Evaluation.} When it comes to evaluating the performance of dialogue emotion generation tasks, relying solely on a small group of volunteers may not provide a comprehensive understanding of the model's capabilities. The preferences of this limited group towards the SOTA model and ChatGPT's generation might not be reflective of a more general audience. Therefore, there is a need for a more objective and universal evaluation method to provide a more accurate assessment of the model's performance.



\bibliography{anthology,custom}
\bibliographystyle{acl_natbib}

\appendix
\section{Definitions of Strategies in ESC}
\label{app:stra}

There are overall 8 types of support strategies that are originally annotated in the \textsc{ESConv} dataset:
\begin{itemize}
    \item \textbf{Question}: ask for information related to the problem to help the help-seeker articulate the issues that they face.  
    \item \textbf{Restatement or Paraphrasing}: 
    a simple, more concise rephrasing of the support-seeker’s statements that could help them see their situation more clearly.
    \item \textbf{Reflection of Feelings}: describe the help-seeker’s feelings to show the understanding of the situation and empathy.
    \item \textbf{Self-disclosure}: share similar experiences or emotions that the supporter has also experienced to express your empathy.
    \item \textbf{Affirmation and Reassurance}: affirm the help-seeker’s ideas, motivations, and strengths to give reassurance and encouragement. 
    \item \textbf{Providing Suggestions}: provide suggestions about how to get over the tough and change the current situation.
    \item \textbf{Information}: provide useful information to the help-seeker, for example with data, facts, opinions, resources, or by answering questions.
    \item \textbf{Others}: other support strategies that do not fall into the above categories.
\end{itemize}

\end{document}